%% file: main.tex
\documentclass{article}
\newcommand{\method}{PreAct-Nav}
\usepackage[T1]{fontenc}
\usepackage{iclr2027_conference,times}
\usepackage{arxiv-preprint}
\usepackage{pifont}
\usepackage{microtype}
\usepackage{amsmath,amssymb,graphicx,booktabs,tabularx}
\usepackage{xcolor,hyperref,url}
\usepackage{placeins}
\usepackage{capt-of}
\hypersetup{
    colorlinks=true,
    linkcolor=red,
    citecolor=gray,
    urlcolor=magenta
}
\input{table_style}
\input{appendix_style}

\newcommand{\R}{\mathcal{R}}
\newcommand{\W}{\mathcal{W}}
\newcommand{\M}{\mathcal{M}}
\newcommand{\C}{\mathcal{C}}
\title{PreAct-Nav: Agentic Reasoning Before Action for Urban Navigation}
\input{authors}
\begin{document}
\maketitle
\begin{abstract}
Urban navigation requires embodied agents to pursue long-horizon goals through local decisions based on egocentric observations. However, existing agentic navigation methods often struggle to translate distant goals into coherent local decisions in large-scale physical environments. Their reliance on linguistic reasoning over transient observations or limited history constrains anticipation of the consequences of actions and future conditions, despite the importance of such foresight for navigating long and complex urban routes. To bridge this gap, we propose PreAct-Nav, an agentic navigation framework that equips frozen policies with anticipatory reasoning for robust urban navigation. Our central idea is to anchor local decisions in persistent medium-horizon subgoals, assess the consequences of predicted actions before execution, and continually update the reasoning context using actual outcomes. At its core, a navigation memory module maintains the active subgoal and relevant experience across decisions, translating distant goals into actionable intermediate objectives. We further introduce a predictive world sandbox that uses an action-conditioned world model (AC-WM) to forecast world dynamics conditioned on candidate movements. A vision-language model (VLM) reasoner interprets these predictions under the current subgoal to retain or revise actions. After execution, real observations are used to assess outcomes, correct inconsistent assumptions, and update memory to continue or reformulate the subgoal. Extensive evaluations demonstrate that the proposed PreAct-Nav improves action selection through memory updates and visual prediction, with more pronounced gains on longer routes and routes with more turns.
\end{abstract}
\input{sections_intro_related}
\input{sections_method}
\input{sections_experiments}
\input{sections_discussion}
\input{references_preprint}

\clearpage
\appendix
\input{appendix}
\end{document}

%% file: appendix_style.tex
\usepackage{listings}
\usepackage{tcolorbox}
\tcbuselibrary{breakable,skins,listings}
\definecolor{PromptNavy}{HTML}{203953}
\definecolor{PromptRule}{HTML}{BECBD5}
\definecolor{PromptTint}{HTML}{F7F9FB}
\lstdefinestyle{navprompt}{
  basicstyle=\fontsize{9}{11.4}\selectfont\ttfamily,
  breaklines=true,breakatwhitespace=false,columns=fullflexible,
  breakautoindent=false,breakindent=0pt,
  keepspaces=true,showstringspaces=false,tabsize=2,
  aboveskip=0pt,belowskip=0pt,
  keywordstyle=\color{PromptNavy},commentstyle=\color{black},
  stringstyle=\color{black}
}
\newtcolorbox{navpromptbox}[1]{
  enhanced,breakable,colback=PromptTint,colframe=PromptRule,
  colbacktitle=PromptTint,coltitle=PromptNavy,
  fonttitle=\small\bfseries,title={#1},
  title after break={#1\enspace\normalfont\small(continued)},
  boxrule=0.45pt,arc=1.2mm,left=7pt,right=7pt,top=6pt,bottom=6pt,
  toptitle=5pt,bottomtitle=5pt,before skip=8pt,after skip=9pt,
  pad at break*=3mm
}

%% file: authors.tex
\newcommand{\preprintauthorblock}{%
  \begin{minipage}{\textwidth}
  \centering
  {\normalsize\bfseries
  Jing Xie\textsuperscript{1,\textdagger}\quad
  Shouwei Ruan\textsuperscript{2,\textdagger}\quad
  Yubin Wang\textsuperscript{3,\textdaggerdbl}\quad
  Yuxiang Zhang\textsuperscript{4}\\[3pt]
  Junwei Yang\textsuperscript{5}\quad
  Songchang Jin\textsuperscript{6}\quad
  Dianxi Shi\textsuperscript{6,*}\par}
  \vspace{7pt}
  {\normalfont\fontsize{8.5}{10}\selectfont
  \textsuperscript{1}MoE Key Lab of Artificial Intelligence, Institute of AI, School of Computer Science,\\
  Shanghai Jiao Tong University, Shanghai, China\\[2pt]
  \textsuperscript{2}Beihang University, Beijing, China\\
  \textsuperscript{3}The Hong Kong University of Science and Technology, Hong Kong, China\\[2pt]
  \textsuperscript{4}Tsinghua University, Beijing, China\quad
  \textsuperscript{5}University of Cambridge, Cambridge, UK\\[2pt]
  \textsuperscript{6}Intelligent Game and Decision Lab (IGDL), Beijing, China\par}
  \end{minipage}%
}
\author{\preprintauthorblock}
\makeatletter
\g@addto@macro\@thanks{%
  \footnotetext[2]{Equal contribution.\quad
    \textsuperscript{\textdaggerdbl}Project leader.}%
  \footnotetext[1]{Corresponding author.}%
}
\makeatother
\hypersetup{
  pdftitle={PreAct-Nav: Agentic Reasoning Before Action for Urban Navigation},
  pdfauthor={Jing Xie, Shouwei Ruan, Yubin Wang, Yuxiang Zhang, Junwei Yang, Songchang Jin, Dianxi Shi},
  pdfsubject={Preprint},
  pdfkeywords={urban navigation, agentic reasoning, navigation memory, world models}
}

%% file: sections_intro_related.tex
\section{Introduction}

\begin{figure}[t]
\centering
\includegraphics[width=1.0\textwidth]{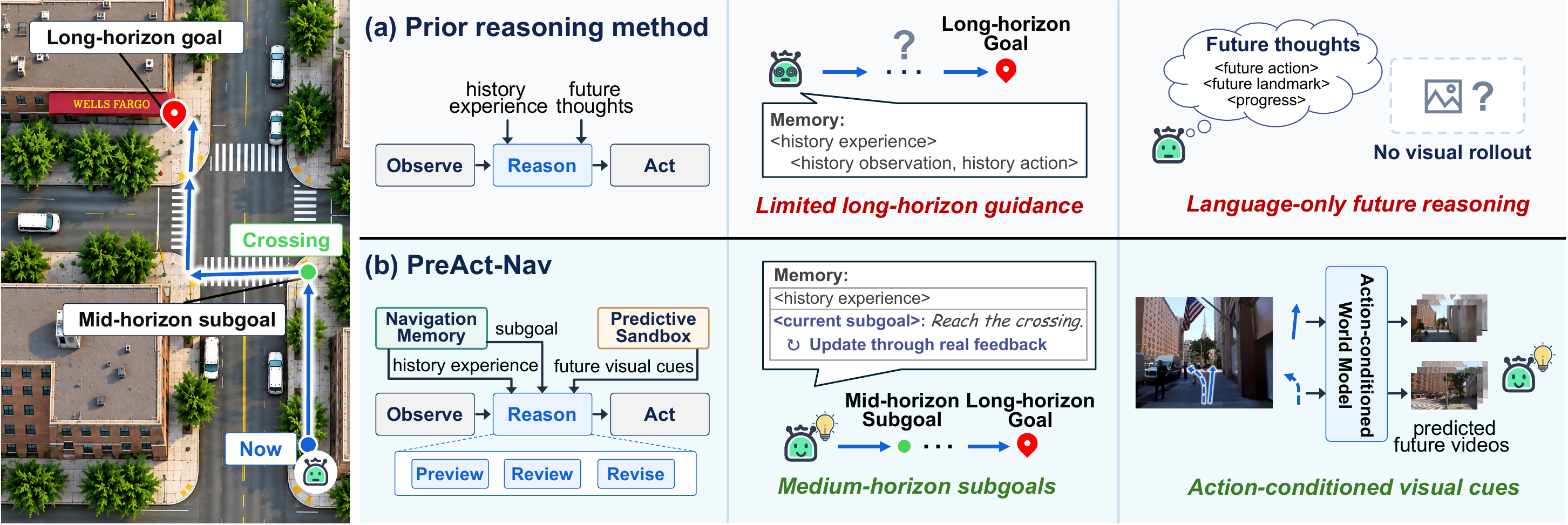}
\par\vspace{-6pt}
\caption{\textbf{Reasoning before action with \method{}.} Navigating across an intersection requires local decisions to serve a long-horizon goal. Compared with reasoning from observed history and language-based future thoughts, \method{} maintains medium-horizon subgoals to decompose long-horizon navigation tasks and uses action-conditioned future visual cues to preview, review, and revise local actions before execution. Real feedback updates memory for subsequent decisions.}
\label{fig:concept}
\par\vspace{-6pt}
\end{figure}

Urban navigation requires embodied agents to reach a destination described in language through a sequence of local movements~\citep{Mei_2026}. The agent sees only part of its surroundings, while the route may extend beyond several intersections and turns. A locally feasible movement is therefore not necessarily useful for reaching the destination. As an example, the route in Figure~\ref{fig:concept} first requires approaching a crossing before continuing toward the goal. Thus, reliable navigation in such environment requires connecting such intermediate steps to the long-horizon goal while anticipating how local actions may affect subsequent progress.

Recent language and vision-language models (VLMs) have enabled navigation agents to reason explicitly about instructions, observations, and possible actions~\citep{zhou2024navgpt,long2024discussnav,ruan2026brain}. Rather than directly executing a policy prediction, these agents use reasoning to assess where to move next. EvoNav~\citep{dai2026evonav} further incorporates historical experience and future thoughts by retrieving past experience and predicting subsequent actions and landmarks. These advances establish reasoning over experience as a practical way to improve navigation decisions.

Despite this progress, connecting local reasoning to long-horizon navigation remains challenging in two respects. \textbf{First}, historical context records past observations and actions, but provides limited guidance for subsequent movements toward the long-horizon goal. Local decisions may therefore be reasonable in isolation yet inconsistent across steps. \textbf{Second}, reasoning about future actions and landmarks in language does not provide visual evidence of what a proposed movement might reveal. A passage that appears plausible from the current view may remain occluded or fail to lead toward the intended route. These limitations make it difficult to determine whether a locally plausible action will support progress toward the destination.

In this paper, we propose \textbf{\method{}}, a training-free agentic reasoning harness that equips frozen navigation policies with anticipatory reasoning. Our central idea is to maintain medium-horizon subgoals to guide local decisions, evaluate proposed actions using predicted visual outcomes before execution, and update navigation memory using execution feedback. A shared VLM reasoner uses the current subgoal, relevant experience, and visual predictions to decide whether to retain or revise the proposed action. This reasoning loop connects action evaluation and revision before execution with memory updates afterward, without updating model weights.

Specifically, the \method{} harness organizes a VLM reasoner and two callable tools into a four-stage reasoning loop (Figure~\ref{fig:architecture}). During \textbf{Action Preview}, the reasoner reads \emph{Navigation Memory}, which maintains the current subgoal and relevant experience across decisions. If the available evidence is insufficient, it queries the \emph{Predictive Sandbox}, which uses an action-conditioned world model (AC-WM) to generate future views for the proposed movement~\citep{Bar_2025_CVPR}. During \textbf{Action Review}, the reasoner uses predicted visual cues to assess whether the proposed action makes progress toward the current subgoal. It retains a supported action or revises a contradicted one; a revised action returns to preview and review within a fixed reasoning budget. In \textbf{Action Execute}, the agent follows the accepted movement. Finally, \textbf{Memory Update} compares the expected cues with the next real observation, records the outcome, and updates subgoal progress for the next decision. By coordinating context, tool use, and execution feedback, the harness maintains medium-horizon guidance across local actions: prediction guides revision, and real feedback updates memory.

We evaluate \method{} on UrbanNav with three navigation policies and two VLM reasoners. Our method achieves relative improvements of 6.71\% in success rate (SR) and 4.19\% in success weighted by path length (SPL) over the benchmark's original policy. Experiments with other policies and reasoners further assess its applicability beyond a single backbone. Component ablations and analyses of individual decisions examine the roles of memory and prediction, with larger prediction gains on longer routes and routes with more turns. In summary, our contributions are:

\ding{182} We introduce \method{}, a training-free navigation harness that reviews and revises policy actions using memory and visual prediction before execution.

\ding{183} We develop Navigation Memory to maintain medium-horizon subgoals alongside observed experience and update their progress through execution feedback.

\ding{184} We expose an action-conditioned world model as a Predictive Sandbox, providing future visual cues for action reasoning under the current subgoal.

\ding{185} We evaluate the harness across navigation policies and VLM reasoners, with component ablations and analyses of memory, prediction, and navigation difficulty.

\vspace{-4pt}
\section{Related Work}

\paragraph{Navigation reasoning.}
Vision-and-language navigation grounds instructions in observed scenes, from navigation graphs to continuous environments~\citep{anderson2018vln,krantz2020vlnce}. Learned policies translate observations and goals into local movements~\citep{Mei_2026,sridhar2024nomad,hirose2026omnivla}, while CityWalker learns urban navigation from web videos~\citep{liu2025citywalker}. LM-Nav composes pretrained models without fine-tuning~\citep{shah2023lmnav}, while CompassNav supervises feasible-action decisions~\citep{li2026compassnav}. Explicit reasoning supports instruction interpretation~\citep{zhou2024navgpt,qiao2025opennav}, multi-expert assessment~\citep{long2024discussnav}, and retrieved experience with future thoughts~\citep{dai2026evonav}. ReAct and Inner Monologue interleave reasoning with feedback~\citep{yao2023react,pmlr-v205-huang23c}, while Toolformer learns when and how to call tools~\citep{schick2023toolformer}. Our frozen reasoner coordinates memory and prediction through a harness to review and revise movements before execution.

\vspace{-3pt}
\paragraph{Agent memory.}
Memory enables agents to reuse experience without encoding every interaction in model weights. Reflexion stores verbal reflections~\citep{NEURIPS2023_1b44b878}, REMem retrieves temporally grounded events~\citep{ICLR2026_5b93eda3}, and M3-Agent builds multimodal episodic and semantic memories~\citep{long2026m3agent}. Agentic Context Engineering incrementally updates contextual playbooks using execution feedback~\citep{zhang2026ace}. In navigation, DUET combines a growing topological map with local observations~\citep{chen2022duet}, and PanoNav retains exploration history to avoid repeated local deadlocks~\citep{Jin_2026}. Our Navigation Memory couples observed experience with an active medium-horizon subgoal that provides shared context for action review and revision. Execution feedback updates its objective and progress cues, so memory tracks both past outcomes and the current purpose of navigation.

\vspace{-3pt}
\paragraph{Action-conditioned world models.}
Video generation provides a visual representation of future plans: UniPi generates goal-conditioned videos and extracts control actions from them~\citep{du2023unipi}. Action-conditioned world models instead predict observations for a specified movement. Navigation World Models use such predictions to plan paths or rank proposed trajectories~\citep{Bar_2025_CVPR}, while Cosmos Policy fine-tunes video models for visuomotor control and planning~\citep{kim2026cosmospolicy}. In computer-use agents, R-WoM grounds language-model simulations in retrieved tutorials~\citep{mei2026rwom}. Access to a world model, however, does not ensure effective use of its predictions~\citep{qian-etal-2026-current}. Our focus is how visual prediction informs navigation reasoning. The Predictive Sandbox exposes the world model as a callable tool; the VLM interprets predicted visual cues against the current subgoal to retain or revise a movement. Real observations subsequently establish progress and update memory, linking predictive reasoning to execution feedback.

%% file: sections_method.tex
\section{Methodology}
\label{sec:method}

\begin{figure}[t]
\centering
\includegraphics[width=\textwidth]{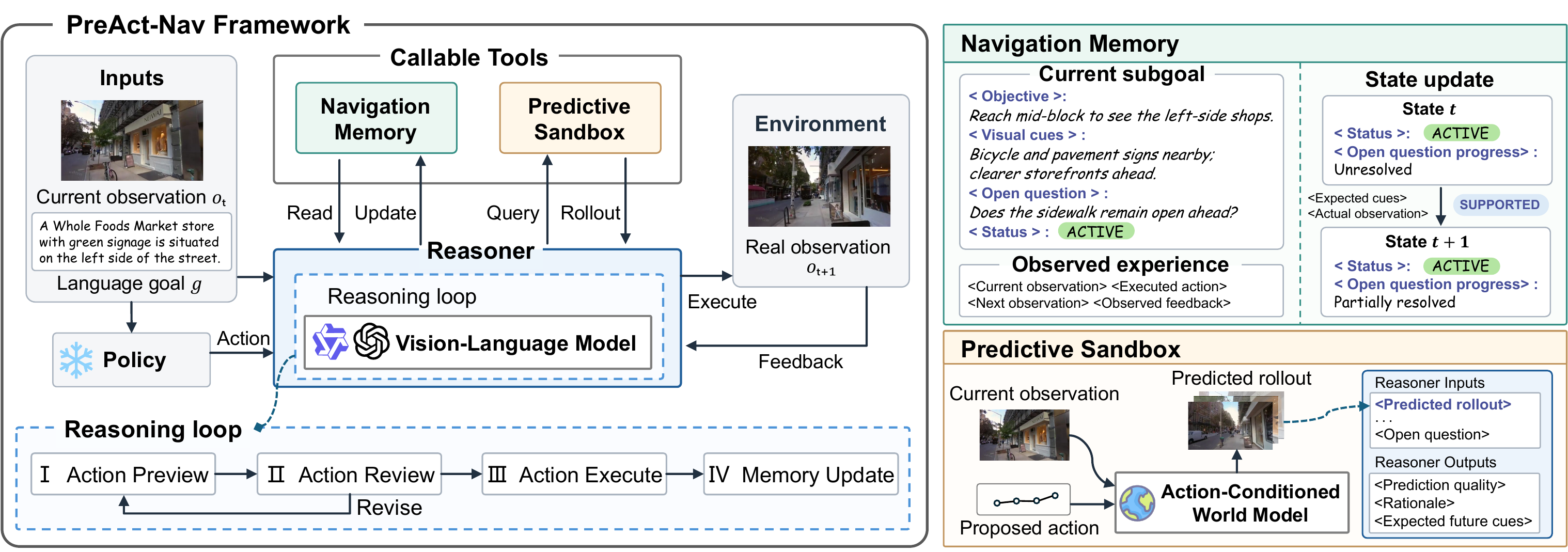}
\par\vspace{-6pt}
\caption{\textbf{Overview of \method{}.} A shared VLM coordinates Navigation Memory and the Predictive Sandbox through Action Preview, Action Review, Action Execute, and Memory Update. Memory provides observed experience and the current subgoal; the sandbox supplies predicted visual cues. Revised actions return to preview. After execution, real observations are checked against expected cues to update subgoal status and open-question progress.}
\label{fig:architecture}
\par\vspace{-6pt}
\end{figure}

\method{} is a training-free agentic reasoning harness for a frozen navigation policy. It organizes a shared VLM reasoner $\R$ and two callable tools into a loop that previews, reviews, and revises actions before execution. \textbf{Navigation Memory} maintains a medium-horizon subgoal and observed experience; the \textbf{Predictive Sandbox} provides action-conditioned future visual cues. Real execution feedback updates memory for subsequent decisions. Figure~\ref{fig:architecture} summarizes the harness, implemented through structured prompts and tool calls with all model weights frozen.

\subsection{Problem formulation}
\label{sec:formulation}

\paragraph{Navigation task and policy actions.}
Given a language description $g$ of the destination view, the agent must reach the destination and recognize arrival from egocentric RGB observations $o_t$. At step $t$, a movement command $u_t$ produces $o_{t+1}$, whereas \textsc{Stop} ends navigation. The frozen policy returns an action and stopping score, $(a_t^0,p_t^{\mathrm{stop}})=\pi(g,\mathcal I_t^\pi,\mathbf p_t)$, using visual inputs $\mathcal I_t^\pi$ and, when required, position and orientation inputs $\mathbf p_t$. Across policies, actions share a waypoint representation $a_t^0=(\mathbf w_{t,1}^0,\ldots,\mathbf w_{t,K}^0)$, where $\mathbf w_{t,k}^0\in\mathbb R^2$ is measured in metres relative to the current position and viewing direction: $x$ points right and $y$ forward.

\vspace{-3pt}
\paragraph{Reasoning context.}
The reasoner receives the destination, current and selected recent observations $\mathcal I_t$, navigation memory $\M_t$, and the policy action:
\begin{equation}
 \C_t=(g,\mathcal I_t,\M_t,a_t^0),\qquad
 \M_t=(E_t,S_t),
\label{eq:context}
\end{equation}
where $E_t$ stores observed experience and $S_t$ represents the current subgoal. Superscript $j=0$ denotes the policy action and $j>0$ its successive revisions. The reasoning history $\mathcal T_t^j=((a_t^i,r_t^i))_{i<j}$ retains earlier actions and reviews, with $\mathcal T_t^0=\varnothing$. Observations and memory remain fixed until execution advances $t$. Subscripts on $\R$ denote role-specific prompts for the same VLM.

\subsection{Navigation memory}
\label{sec:memory}

Navigation Memory maintains medium-horizon subgoals to decompose long-horizon tasks. Experience $E_t$ records past outcomes; the current subgoal $S_t$ guides action review and revision.

\vspace{-3pt}
\paragraph{Observed experience.}
Each record $e_i=(o_i,a_i,o_{i+1},v_{i+1})$ contains \emph{Current observation}, \emph{Executed action}, \emph{Next observation}, and \emph{Observed feedback}, as shown in Figure~\ref{fig:architecture}. Recent images and concise summaries are retained in temporal order to identify actions that failed to produce expected progress. Past actions use the local coordinate frame of their associated observations.

\vspace{-3pt}
\paragraph{Current subgoal.}
A subgoal specifies an intermediate objective across local actions:
\begin{equation}
 S_t=(z_t,c_t,q_t,\sigma_t),
\label{eq:memory}
\end{equation}
with four fields: \emph{Objective} $z_t$ states what to accomplish; \emph{Visual cues} $c_t$ identify nearby visual anchors and signs of progress or completion; \emph{Open question} $q_t$ records an unresolved route question; and \emph{Status} $\sigma_t$ indicates whether the subgoal is \textsc{Active}, \textsc{Achieved}, or \textsc{Contradicted}. The reasoner establishes the subgoal from the destination, current observations, and retained experience.

For example, Figure~\ref{fig:architecture} specifies reaching mid-block to see the left-side shops, with clearer storefronts indicating progress and sidewalk continuity remaining an open question. The subgoal conditions the reasoner; the policy remains conditioned on the final destination $g$.

\vspace{-3pt}
\paragraph{State update from execution feedback.}
After executing $a_t$, the reasoner compares the next real observation with the \emph{Expected future cues} $\nu_t$ retained from action review:
\begin{equation}
 v_{t+1}=\R_{\mathrm{update}}(g,S_t,a_t,\nu_t,o_t,o_{t+1}).
\label{eq:feedback}
\end{equation}
Feedback $v_{t+1}$ reports \textsc{Supported}, \textsc{Achieved}, \textsc{Contradicted}, or \textsc{Unresolved}, with a brief explanation of open-question progress. Support indicates compatible progress, achievement requires observed completion cues, contradiction calls for reconsideration, and unresolved feedback preserves the open question. Supported progress can leave a subgoal active with a partially resolved open question (Figure~\ref{fig:architecture}).

The observed transition is appended to experience, and the feedback updates the subgoal:
\begin{equation}
 \begin{aligned}
 E_{t+1}&=\operatorname{Select}\bigl(E_t\mathbin{\Vert}(e_t)\bigr),\\
 S_{t+1}&=\operatorname{Update}(S_t,v_{t+1}),\qquad
 \M_{t+1}=(E_{t+1},S_{t+1}).
 \end{aligned}
\label{eq:subgoal-transition}
\end{equation}
Here $\Vert$ appends a record and $\operatorname{Select}$ retains recent experience. Supported or unresolved feedback preserves the active objective; achievement or contradiction closes it before a new objective is established. Only real transitions enter observed experience. Retention settings appear in Appendix~\ref{app:implementation}.

\subsection{Predictive sandbox}
\label{sec:prediction}

The Predictive Sandbox exposes an action-conditioned world model $\W$ as a callable tool. From the current observation and proposed action, it generates a short video, or \emph{predicted rollout}, for review under the current subgoal.

\vspace{-3pt}
\paragraph{Action-conditioned prediction.}
Let $\mathbf C(a_t^j)$ encode the proposed movement as the positions and viewing directions required by the world model. The tool query and returned rollout are
\begin{equation}
 Q_t^j=(o_t,\mathbf C(a_t^j),\ell,\epsilon),\qquad
 \widehat O_t^j=\W(Q_t^j)
       =(\widehat o_{t,1}^j,\ldots,\widehat o_{t,L}^j),
\label{eq:prediction}
\end{equation}
where $\ell$ is a scene-continuation prompt, $\epsilon$ controls generation randomness, and $L$ counts predicted frames. Across revisions, the initial observation, prompt, and seed remain fixed; only the proposed movement changes. Appendix~\ref{app:implementation} specifies action encoding and generation settings.

\vspace{-3pt}
\paragraph{Visual evidence collection.}
The reasoner inspects the rollout and its action using the destination, observations, subgoal, and open question. It returns three fields (Figure~\ref{fig:architecture}): \emph{Prediction quality} assesses coherence, movement consistency, and relevance; \emph{Rationale} links visual cues to the action assessment; and \emph{Expected future cues} specify changes to check after execution.

The reasoner looks for progress toward the subgoal, not completion within a single rollout. Incoherent or inconclusive predictions defer to observed evidence. Tool records retain the conditioning action, rollout, and assessment for revision within the current decision.

\subsection{Agentic reasoning harness}
\label{sec:decision}

The harness organizes context, tool use, and execution feedback around the VLM's judgments and revisions. Its four-stage loop connects memory retrieval and on-demand prediction to action review, execution, and memory update. Each decision permits at most $B$ reviewed actions, including the policy action, and $B-1$ revisions. The reasoning budget $B\geq1$ differs from the waypoint count $K$.

\vspace{-3pt}
\paragraph{Stage I: Action Preview.}
Starting with $a_t^0$, the reasoner performs an observation-based review $\bar r_t^j=\R_{\mathrm{review}}(\C_t,\mathcal T_t^j,a_t^j,\varnothing)$ using the interface defined below. It requests a preview when the action assessment is unresolved and a forecast could supply missing evidence:
\begin{equation}
 \lambda_t^j=\mathbf 1\!\left[
 \bar\omega_t^j=\textsc{Unresolved}\ \land\ \bar\kappa_t^j=1
 \right],
\label{eq:query}
\end{equation}
where $\bar\omega_t^j$ is the observation-based action assessment and $\bar\kappa_t^j$ indicates whether a preview would help resolve the open question. The available prediction is
\begin{equation}
 \widehat O_t^j=
 \begin{cases}
 \W(Q_t^j),&\lambda_t^j=1,\\
 \varnothing,&\lambda_t^j=0.
 \end{cases}
\label{eq:preview}
\end{equation}
Each action receives at most one preview; otherwise, the observation-based assessment is retained.

\vspace{-3pt}
\paragraph{Stage II: Action Review.}
The reasoner reviews the proposed action against the current subgoal, incorporating the predicted rollout when available:
\begin{equation}
 r_t^j=(\omega_t^j,\chi_t^j,\rho_t^j,\nu_t^j,\kappa_t^j)
 =\R_{\mathrm{review}}(\C_t,\mathcal T_t^j,a_t^j,\widehat O_t^j).
\label{eq:assessment}
\end{equation}
The action assessment $\omega_t^j$ is \textsc{Supported}, \textsc{Contradicted}, or \textsc{Unresolved}. The remaining entries are prediction quality $\chi_t^j$, rationale $\rho_t^j$, expected future cues $\nu_t^j$, and preview usefulness $\kappa_t^j$. Without a rollout, prediction quality is not assessed; other fields use observations and memory.

A supported assessment accepts the action. A contradicted assessment identifies a reason to change it, such as a movement toward a blocked passage. If the reasoning budget permits, the same VLM generates a revised action with its rationale and expected cues:
\begin{equation}
 \begin{aligned}
 (a_t^{j+1},\rho_t^{j+1},\nu_t^{j+1})
 &=\R_{\mathrm{revise}}(\C_t,\mathcal T_t^j,a_t^j,r_t^j,\widehat O_t^j),\\
 &\text{if }\omega_t^j=\textsc{Contradicted}\ \land\ j+1<B.
 \end{aligned}
\label{eq:revision}
\end{equation}
The revision addresses the identified problem under the subgoal and available evidence. The harness appends $(a_t^j,r_t^j)$ to reasoning history. A valid revision must satisfy the action-format and movement checks in Appendix~\ref{app:implementation} and differ from previous proposals. Writing validity as $V_t^{j+1}=1$, with $V_t^0=1$, a valid action returns to Stage I at $j\leftarrow j+1$.

\vspace{-3pt}
\paragraph{Stage III: Action Execute.}
The agent executes the first supported action. The inner loop also terminates when the budget is exhausted, a revision is invalid or repeated, or the available evidence leaves the assessment unresolved. If no action is supported, it retains the policy action. Let $n_t\leq B$ be the number of reviewed actions and $\mathcal J_t=\{j:0\leq j<n_t,\ V_t^j=1,\ \omega_t^j=\textsc{Supported}\}$. The executed movement is
\begin{equation}
 j_t=
 \begin{cases}
 \min\mathcal J_t,&\mathcal J_t\neq\varnothing,\\
 0,&\text{otherwise},
 \end{cases}
 \qquad a_t=a_t^{j_t}.
\label{eq:decision}
\end{equation}
The selected action's assessment and expected cues are retained for memory update. Without intervention, the agent follows the policy action.

\vspace{-3pt}
\paragraph{Stage IV: Memory Update.}
The reasoner compares the next real observation with the executed action's expected cues using Equation~\ref{eq:feedback}. Equation~\ref{eq:subgoal-transition} updates experience and subgoal progress, yielding $\M_{t+1}$ for context $\C_{t+1}$. Prediction guides revision; real feedback establishes progress.

%% file: sections_experiments.tex
\section{Experiments}
\label{sec:experiments}

We evaluate \method{} on UrbanNav~\citep{Mei_2026} and CityWalker~\citep{liu2025citywalker}, examine transfer across frozen policies, and ablate Navigation Memory and the Predictive Sandbox. Qualitative examples illustrate action revision and memory update. Appendices~\ref{app:analysis} and~\ref{app:cost} report route-level analyses, reasoning diagnostics, and inference costs.

\subsection{Experimental Setup}
\label{sec:setup}

\paragraph{Tasks and evaluation.}
UrbanNav~\citep{Mei_2026} provides first-person urban trajectories and language descriptions of destination views. We evaluate on its test-seen and test-unseen splits, with 4,179 and 4,634 episodes after filtering watermark-related descriptions. On CityWalker~\citep{liu2025citywalker}, we use 154 routes drawn from nine recordings, with each destination specified by its position relative to the agent. Both tasks use a three-metre success radius. UrbanNav allows 25 decisions per episode and CityWalker allows 100. Methods share the evaluation protocol within each task; Appendix~\ref{app:implementation} gives its action and observation rules.

\vspace{-3pt}
\paragraph{Evaluation metrics.}
We report success rate (SR), success weighted by path length (SPL)~\citep{anderson2018navigationeval}, navigation error (NE), normalized dynamic time warping (nDTW), success weighted by nDTW (SDTW)~\citep{ilharco2019dtw}, and oracle success rate (OSR). SR and SPL measure task completion and path efficiency; the remaining metrics measure final goal distance, reference-path alignment, and whether the goal region is reached at any step. NE is in metres and all other metrics are percentages. Improvements are percentage-point differences computed before rounding. Metric definitions are in Appendix~\ref{app:metrics}.

\vspace{-3pt}
\paragraph{Models and baselines.}
The default configuration combines the UrbanNav policy~\citep{Mei_2026}, Qwen3.8-27B~\citep{qwen2026qwen38}, and the official Cosmos3-Nano checkpoint~\citep{nvidia2026cosmos3nano} as the action-conditioned world model. We also evaluate GPT-5.6 Sol~\citep{openai2026gpt56} as the reasoner and NoMaD~\citep{sridhar2024nomad} and OmniVLA~\citep{hirose2026omnivla} as navigation policies. Baselines include the policy alone, a Direct VLM, and NavGPT~\citep{zhou2024navgpt}, Open-Nav~\citep{qiao2025opennav}, DiscussNav~\citep{long2024discussnav}, and EvoNav~\citep{dai2026evonav}. All reasoning baselines are evaluated with Qwen3.8-27B; NavGPT and Open-Nav are additionally evaluated with GPT-5.6 Sol. The reasoning baselines receive three observed RGB frames, the policy proposal, and policy-centred candidate viewpoints, and are evaluated under the same protocol.

\vspace{-3pt}
\paragraph{Implementation details.}
All pretrained model weights remain frozen during evaluation. Navigation policies share the harness interfaces for memory, prediction, and action review. Appendices~\ref{app:algorithm}, \ref{app:prompts}, and~\ref{app:implementation} describe the reasoning loop, prompts, and video generation.

\subsection{Main Results}
\label{sec:main-results}
\input{provenance/table_main}

\paragraph{Comparison on UrbanNav.}
Table~\ref{tab:main} compares \method{} with the original policy and reasoning baselines on test-seen and test-unseen. On test-unseen, Qwen raises SR from 78.14\% to 83.38\% and SPL from 74.73\% to 77.87\%, while reducing NE from 5.9659 to 5.3353 m. It also improves nDTW, SDTW, and OSR, indicating better reference-path alignment and more trajectories reaching the goal region. Relative to Open-Nav, the strongest reasoning baseline in test-unseen SR, the gain is 4.14 points. On test-seen, Qwen reaches 85.09\% SR and 82.79\% SPL, improving over the original policy by 2.30 and 1.63 points. Both splits improve on all six metrics, with a larger SR gain on test-unseen.

\vspace{-3pt}
\paragraph{Different VLM reasoners.}
With GPT-5.6 Sol, \method{} reaches 84.18\% SR and 80.57\% SPL on test-unseen, and 85.91\% SR and 84.19\% SPL on test-seen. With the same GPT reasoner, Open-Nav reaches 78.85\% SR on test-unseen and NavGPT reaches 83.32\% on test-seen; \method{} exceeds these strongest baselines within the GPT group by 5.33 and 2.58 SR points, respectively. Completion and efficiency improve with both an open-weight and a proprietary VLM.

\input{provenance/table_citywalker}
\paragraph{Comparison on CityWalker.}
Table~\ref{tab:citywalker} compares reasoning methods on the frozen CityWalker policy. With Qwen as the shared reasoner, \method{} achieves 79.22\% SR and 78.94\% SPL, exceeding NavGPT, the strongest listed baseline in SR, by 5.84 and 8.74 points. It also achieves the lowest NE and highest nDTW, SDTW, and OSR, improving completion, path efficiency, and reference-path alignment. All methods share the CityWalker evaluation protocol.

\subsection{Policy Transfer}
\label{sec:policy-transfer}
\input{provenance/table_transfer}
Table~\ref{tab:transfer} evaluates the harness with three frozen navigation policies on test-unseen. Beyond the UrbanNav policy, NoMaD gains 2.78 SR points and 2.76 SPL points, while OmniVLA gains 3.58 and 3.39 points. SDTW increases from 73.55\% to 76.13\% for NoMaD and from 79.98\% to 83.33\% for OmniVLA. OmniVLA improves on all six metrics; for NoMaD, OSR is unchanged and NE and nDTW remain nearly constant. NoMaD's gain primarily reflects better completion and path efficiency. These results show that the harness improves navigation without changing policy weights.

\subsection{Ablation Study}
\label{sec:ablation}
\input{provenance/ablation_with_routes}

\paragraph{Component ablation.}
Table~\ref{tab:ablation} evaluates the Reasoning Harness and its two callable tools: Navigation Memory Tool and Predictive Sandbox Tool. Without either tool, the harness reviews and revises policy actions, reaching 80.56\% SR and 76.74\% SPL. Adding the Navigation Memory Tool raises SR to 82.37\%, while adding only the Predictive Sandbox Tool yields 81.48\%. The full method achieves the highest SR of 83.38\%, the lowest NE, and the highest nDTW, SDTW, and OSR. Its SPL is also highest at 77.87\%, versus 77.24\% with the sandbox alone.

\vspace{-3pt}
\paragraph{Route length and turning.}
Figure~\ref{fig:route-gains} reports gains over the UrbanNav policy within each route-length and turning interval. The full method improves SR and reduces NE in every interval. Its SR gain reaches 9.60 points on 30--50 m routes and 7.92 points on routes with 360--540 degrees of cumulative turning. The largest NE reductions occur in the longest and most turning-intensive groups, although SR gains decrease in these extreme groups. The Predictive Sandbox Tool generally improves on the harness with the Navigation Memory Tool alone, with larger gains on longer routes and routes with greater cumulative turning. The curves show point estimates; group sizes and the uncertainty protocol are given in the Appendix~\ref{app:analysis}.

\subsection{Qualitative Results}
\label{sec:qualitative}

\begin{figure}[t]
\centering
\includegraphics[width=\textwidth]{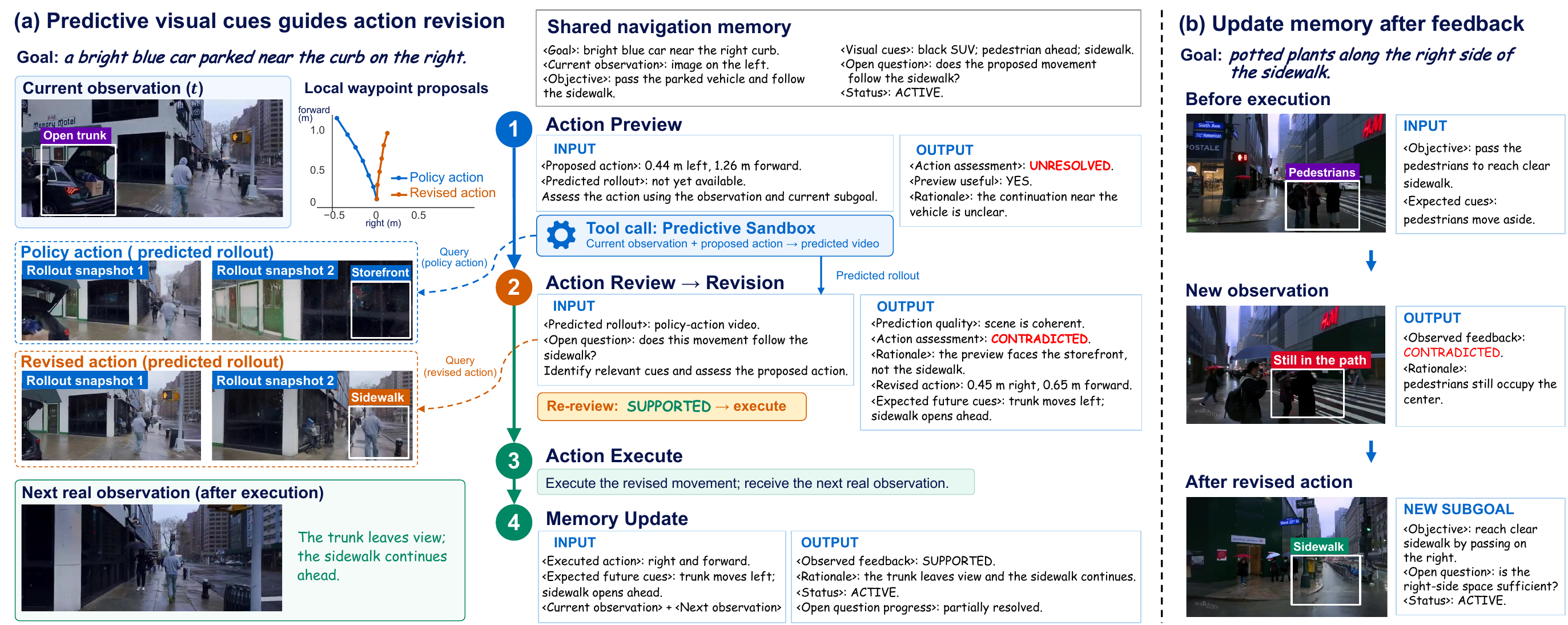}
\par\vspace{-6pt}
\caption{\textbf{Action revision and memory update.} (a) Predicted cues prompt a revision from the storefront toward the sidewalk, supported by the next real observation. (b) Pedestrians remain in the passage, prompting a new subgoal to pass on the right.}
\label{fig:traces}
\par\vspace{-6pt}
\end{figure}

Figure~\ref{fig:traces} illustrates how predicted visual cues and real feedback support the reasoning loop. In panel (a), the policy-action preview turns toward a storefront rather than following the sidewalk. This cue motivates a rightward revision, whose preview shows the sidewalk continuing ahead. After execution, the next observation supports the expected passage, and memory records progress under the current subgoal. Panel (b) illustrates the complementary case: pedestrians remain in the passage instead of moving aside, contradicting the expected cues. Memory updates the objective to passing on the right and retains the open question of whether that space is sufficient. The examples connect action preview and review with execution and subsequent memory update.

\textbf{We also present a qualitative real-robot navigation case in an outdoor plaza (Appendix~\ref{app:real-world}).} The example illustrates how an exterior-route subgoal connects action review and revision with observation-based memory update.

\FloatBarrier

%% file: provenance/table_main.tex
\begin{table}[t]
\vspace{-3pt}
\centering
\caption{\textbf{Comparison on UrbanNav.} All methods use the UrbanNav policy, with rows grouped by VLM reasoner. NE is in metres; other metrics are percentages. Best results are bold.}
\label{tab:main}
\label{tab:main-seen}
\navtableformat
\fontsize{8}{10}\selectfont
\setlength{\tabcolsep}{1pt}
\newcommand{\splitmetrichead}{%
  \mbox{\textbf{SR}$\uparrow$} &
  \mbox{\textbf{SPL}$\uparrow$} &
  \mbox{NE$\downarrow$} &
  \mbox{nDTW$\uparrow$} &
  \mbox{SDTW$\uparrow$} &
  \mbox{OSR$\uparrow$}}
\begin{tabular*}{\textwidth}{@{\extracolsep{\fill}}l*{6}{r}@{\hspace{5pt}}*{6}{r}@{}}
\toprule
& \multicolumn{6}{c}{Test-seen} & \multicolumn{6}{c}{Test-unseen}\\
\cmidrule(lr){2-7}\cmidrule(l){8-13}
Method & \splitmetrichead & \splitmetrichead\\
\midrule
UrbanNav & 82.79 & 81.16 & 3.3596 & 94.10 & 81.73 & 91.91 & 78.14 & 74.73 & 5.9659 & 87.73 & 76.13 & 84.92\\
\rowcolor{reasonergray}
\multicolumn{13}{l}{\textit{Reasoner: Qwen3.8-27B}}\\
Direct VLM & 83.13 & 80.41 & 3.3192 & 94.31 & 82.02 & 92.49 & 79.05 & 75.58 & 5.9187 & 88.02 & 76.99 & 85.63\\
NavGPT & 83.23 & 80.54 & 3.3090 & 94.35 & 82.12 & 92.49 & 79.11 & 75.65 & 5.9017 & 88.09 & 77.06 & 85.67\\
Open-Nav & 83.15 & 80.45 & 3.3175 & 94.31 & 82.04 & 92.44 & 79.24 & 75.77 & 5.8964 & 88.11 & 77.18 & 85.74\\
DiscussNav & 83.08 & 80.47 & 3.3311 & 94.27 & 81.98 & 92.39 & 78.90 & 75.00 & 5.8659 & 88.19 & 76.94 & 85.69\\
EvoNav & 83.25 & 80.61 & 3.2892 & 94.38 & 82.16 & 92.51 & 78.87 & 74.59 & 5.8376 & 88.16 & 76.91 & 85.58\\
\midrule
\textbf{Ours} & 85.09 & 82.79 & 2.8690 & 96.27 & 84.21 & 96.51 & 83.38 & 77.87 & 5.3353 & 90.47 & 81.30 & 90.12\\
\rowcolor{reasonergray}
\multicolumn{13}{l}{\textit{Reasoner: GPT-5.6 Sol}}\\
NavGPT & 83.32 & 81.00 & 3.2878 & 94.38 & 82.23 & 92.49 & 78.55 & 74.56 & 5.8943 & 88.00 & 76.60 & 85.30\\
Open-Nav & 83.13 & 80.67 & 3.3143 & 94.29 & 82.02 & 92.41 & 78.85 & 74.70 & 5.8785 & 88.19 & 76.89 & 85.65\\
\midrule
\textbf{Ours} & \textbf{85.91} & \textbf{84.19} & \textbf{2.8075} & \textbf{96.46} & \textbf{85.07} & \textbf{96.77} & \textbf{84.18} & \textbf{80.57} & \textbf{5.2093} & \textbf{90.90} & \textbf{82.26} & \textbf{90.33}\\
\bottomrule
\end{tabular*}
\par\vspace{-6pt}
\end{table}

%% file: provenance/table_citywalker.tex
\begin{table}[t]
\vspace{-3pt}
\centering
\caption{\textbf{Comparison on CityWalker.} All methods use the frozen CityWalker policy and Qwen reasoner. NE is in metres; other metrics are percentages. Baseline adaptations and evaluation settings are detailed in Appendix~\ref{app:citywalker}.}
\label{tab:citywalker}
\navtableformat
\begin{tabular*}{\textwidth}{@{\extracolsep{\fill}}lrrrrrr@{}}
\toprule
Method & \navmetrichead\\
\midrule
Direct VLM & 43.51 & 43.11 & 18.28 & 37.44 & 32.33 & 43.51\\
NavGPT & 73.38 & 70.19 & 10.12 & 51.00 & 48.32 & 73.38\\
Open-Nav & 64.29 & 60.98 & 12.21 & 45.08 & 41.41 & 64.29\\
EvoNav & 63.64 & 61.05 & 11.97 & 46.45 & 42.34 & 63.64\\
\midrule
\textbf{Ours} & \textbf{79.22} & \textbf{78.94} & \textbf{8.31} & \textbf{68.77} & \textbf{65.85} & \textbf{79.22}\\
\bottomrule
\end{tabular*}
\par\vspace{-6pt}
\end{table}

%% file: provenance/table_transfer.tex
\begin{table}[t]
\vspace{-3pt}
\centering
\caption{\textbf{Comparison across navigation policies on test-unseen.} All pairs use Qwen. Green parentheses show changes from the corresponding frozen policy: percentage points for rate metrics and metres for NE. Changes are computed before rounding.}
\label{tab:transfer}
\navtableformat
\fontsize{8.5}{10.5}\selectfont
\setlength{\tabcolsep}{2pt}
\begin{tabular*}{\textwidth}{@{\extracolsep{\fill}}l*{6}{r@{\hspace{1pt}}l}@{}}
\toprule
Policy & \textbf{SR}$\uparrow$ & & \textbf{SPL}$\uparrow$ & & NE$\downarrow$ & & nDTW$\uparrow$ & & SDTW$\uparrow$ & & OSR$\uparrow$ & \\
\midrule
UrbanNav & 78.14 &  & 74.73 &  & 5.9659 &  & 87.73 &  & 76.13 &  & 84.92 & \\
\textbf{+ harness} & 83.38 & \navdelta{+5.24} & 77.87 & \navdelta{+3.13} & 5.3353 & \navdelta{-0.6306} & 90.47 & \navdelta{+2.74} & 81.30 & \navdelta{+5.17} & 90.12 & \navdelta{+5.20}\\
\addlinespace[2pt]
NoMaD & 76.33 &  & 69.05 &  & 7.2776 &  & 82.73 &  & 73.55 &  & 81.20 & \\
\textbf{+ harness} & 79.11 & \navdelta{+2.78} & 71.81 & \navdelta{+2.76} & 7.2727 & \navdelta{-0.0050} & 82.74 & \navdelta{+0.01} & 76.13 & \navdelta{+2.58} & 81.20 & \navdelta{0.00}\\
\addlinespace[2pt]
OmniVLA & 82.65 &  & 78.06 &  & 5.8666 &  & 87.77 &  & 79.98 &  & 86.90 & \\
\textbf{+ harness} & 86.23 & \navdelta{+3.58} & 81.44 & \navdelta{+3.39} & 5.7795 & \navdelta{-0.0871} & 88.36 & \navdelta{+0.58} & 83.33 & \navdelta{+3.35} & 88.07 & \navdelta{+1.17}\\
\bottomrule
\end{tabular*}
\par\vspace{-6pt}
\end{table}

%% file: provenance/ablation_with_routes.tex
\begin{table}[t]
\vspace{-3pt}
\begingroup
\input{provenance/table_ablation_body}
\endgroup
\par\vspace{4pt}
\centering
\includegraphics[width=\textwidth]{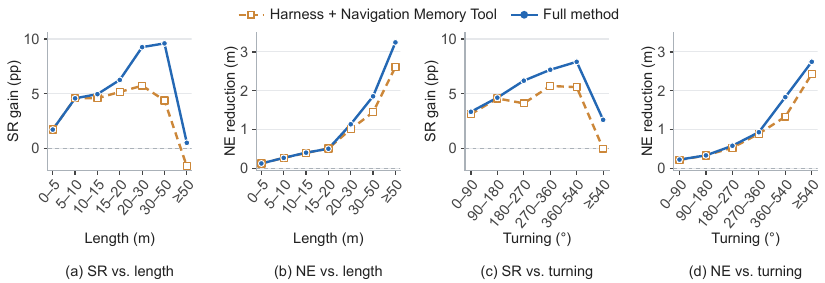}
\par\vspace{-6pt}
\captionof{figure}{\textbf{Gains by route length and turning.} SR gain (pp) and NE reduction (m) over the UrbanNav policy, grouped by (a,b) reference-route length and (c,d) cumulative turning. Both variants use the Reasoning Harness and Navigation Memory Tool; the full method also uses the Predictive Sandbox Tool. Positive values indicate improvement.}
\label{fig:route-gains}
\par\vspace{-6pt}
\end{table}

%% file: provenance/table_ablation_body.tex
\centering
\caption{\textbf{Component ablation on test-unseen.} $\checkmark$/$\times$ indicate enabled/disabled components. The first row is the UrbanNav policy. Without either tool, the Reasoning Harness reviews and revises policy actions. NE is in metres; other metrics are percentages.}
\label{tab:ablation}
\navtableformat
\fontsize{8.5}{10.5}\selectfont
\setlength{\tabcolsep}{2pt}
\begin{tabular*}{\textwidth}{@{\extracolsep{\fill}}cccrrrrrr@{}}
\toprule
\shortstack{Reasoning\\Harness} & \shortstack{Navigation\\Memory Tool} & \shortstack{Predictive\\Sandbox Tool} & \navmetrichead\\
\midrule
$\times$ & $\times$ & $\times$ & 78.14 & 74.73 & 5.9659 & 87.73 & 76.13 & 84.92\\
\midrule
$\checkmark$ & $\times$ & $\times$ & 80.56 & 76.74 & 5.6410 & 89.60 & 78.65 & 88.37\\
$\checkmark$ & $\checkmark$ & $\times$ & 82.37 & 76.91 & 5.3986 & 90.34 & 80.34 & 89.90\\
$\checkmark$ & $\times$ & $\checkmark$ & 81.48 & 77.24 & 5.4791 & 89.87 & 79.47 & 88.89\\
$\checkmark$ & $\checkmark$ & $\checkmark$ & \textbf{83.38} & \textbf{77.87} & \textbf{5.3353} & \textbf{90.47} & \textbf{81.30} & \textbf{90.12}\\
\bottomrule
\end{tabular*}

%% file: sections_discussion.tex
\vspace{-4pt}
\section{Conclusion and Limitations}

We introduced \method{}, a training-free agentic reasoning harness that reviews and revises navigation actions before execution. Navigation Memory maintains medium-horizon subgoals to decompose long-horizon tasks, while the Predictive Sandbox provides action-conditioned future visual cues. A shared VLM uses this context to retain or revise policy actions, and real feedback updates memory. Experiments across frozen policies show improved navigation success, with greater prediction gains on longer routes and routes with more turns.

The harness remains limited by visual prediction quality and cost. Generated videos can provide ambiguous or inconsistent cues, and video generation dominates inference cost. More reliable, faster predictions are needed to extend this loop to time-sensitive navigation.

%% file: references_preprint.tex

%% file: appendix.tex
\raggedbottom
\setcounter{topnumber}{3}
\section*{Supplementary Material}
The appendix opens with the workflow in Appendix~\ref{app:algorithm}, followed by the real-world navigation case in Appendix~\ref{app:real-world}. Appendix~\ref{app:prompts} provides complete prompt templates and their correspondence to the figures; Appendix~\ref{app:implementation} documents the implementation and evaluation protocols. Metrics, supplementary analyses, and inference cost appear in Appendices~\ref{app:metrics}--\ref{app:cost}.

\section{Navigation Workflow}
\label{app:algorithm}

\begin{navpromptbox}{The four-stage navigation loop}
\small
\textbf{Inputs:} destination-view description $g$, local policy $\pi$, shared reasoner $\R$, world model $\W$, and operating settings $\theta$.\\
Initialize observed experience $E$, subgoal context $S$, and pending action cues.\\
\textbf{For each} observation $o_t$ within the navigation budget:\\
\hspace*{1em}If an executed decision is pending, compare its expected cues with $o_t$.\\
\hspace*{2em}Record the observed outcome and update $\M_t=(E_t,S_t)$.\\
\hspace*{1em}Obtain policy action $a_t^0$ and STOP score.\\
\hspace*{1em}Read memory; establish a subgoal if no active objective remains.\\
\hspace*{1em}Initialize the selected movement to the policy proposal.\\
\hspace*{1em}Hold observed context and memory fixed within this decision.\\
\hspace*{1em}For action index $j=0,\ldots,B-1$:\\
\hspace*{2em}Review the current action using observations and memory.\\
\hspace*{2em}If evidence is unresolved and prediction is useful, preview and review it.\\
\hspace*{2em}If supported, select this action and exit the inner loop.\\
\hspace*{2em}If contradicted and budget remains, construct a valid, distinct revision.\\
\hspace*{2em}Continue with that revision; otherwise exit with the policy fallback.\\
\hspace*{1em}Retain the movement to execute and its expected visual cues.\\
\hspace*{1em}Execute the movement and receive the next observation.\\
\textbf{Return:} the executed trajectory and terminal outcome.
\end{navpromptbox}

The procedure summarizes Section~\ref{sec:decision}. Environment observations update experience memory, while generated videos remain associated with their prediction calls.

\section{Real-World Navigation Demonstration}
\label{app:real-world}

\begin{figure}[t]
\centering
\includegraphics[width=\textwidth]{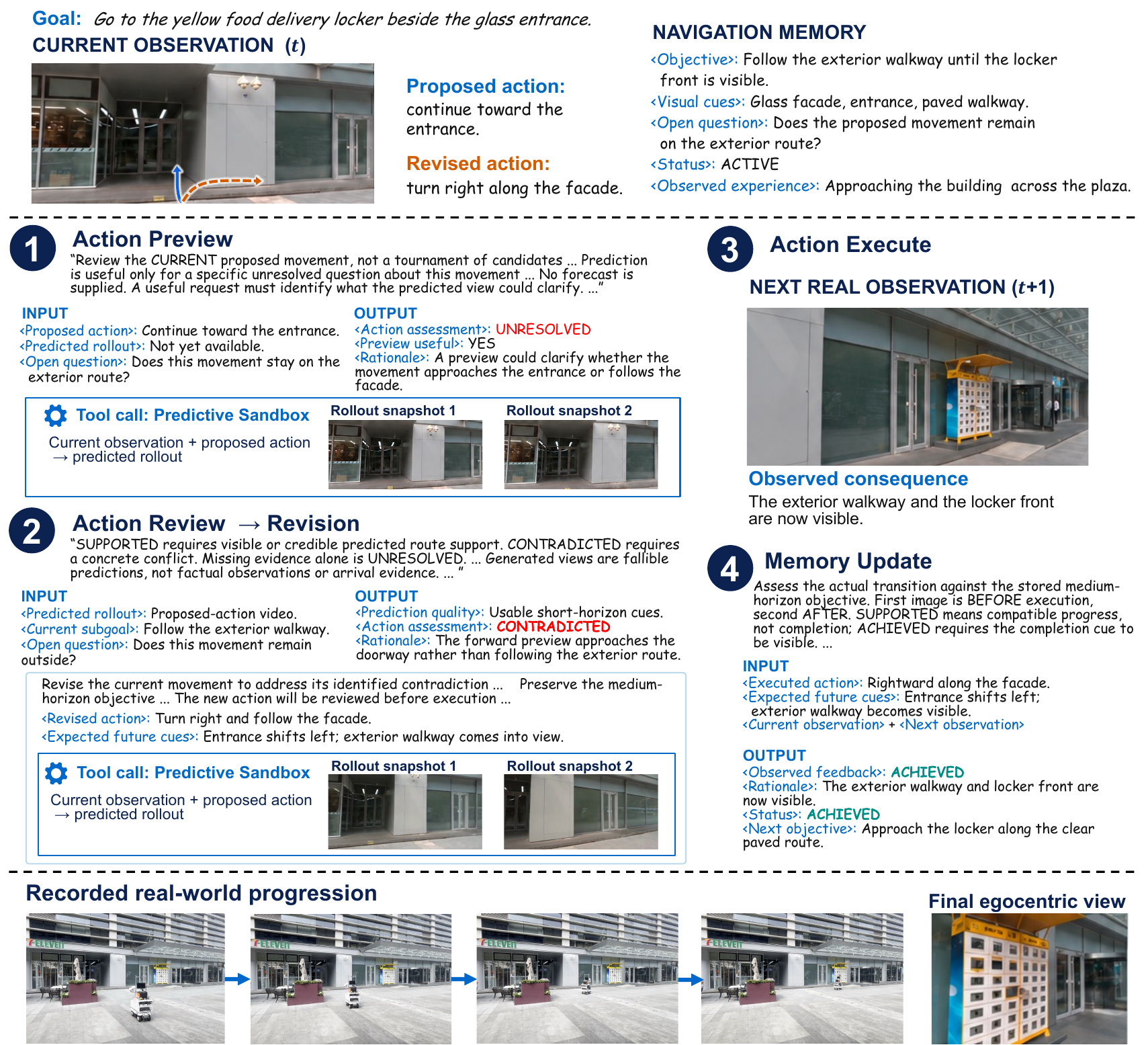}
\caption{\textbf{Real-world navigation case.} A robot follows the building exterior toward a delivery locker. The panels connect the current subgoal, visual previews, action revision, and memory update. }
\label{fig:real-world}
\end{figure}

Figure~\ref{fig:real-world} presents an online closed-loop run of PreAct-Nav on a wheeled robot navigating toward a yellow food-delivery locker beside a glass entrance. During the run, the harness reviews the policy proposal using the current camera observation and Navigation Memory, querying the Predictive Sandbox when additional visual evidence is needed. The selected movement is executed by the robot, and the resulting camera observation updates memory for the next decision. Predicted rollouts are generated online before action execution, whereas the current and next observations are real camera images acquired during physical navigation.
In the illustrated decision, the active subgoal is to follow the exterior walkway until the locker front becomes visible. The policy-action preview approaches the doorway rather than following the exterior route, prompting a rightward revision along the facade. After execution, the next real observation reveals the exterior walkway and locker front. Memory marks the intermediate objective as achieved and establishes the next objective of approaching the locker along the clear paved route. This transition distinguishes subgoal completion from final arrival.

\input{appendix_prompts}
\section{Implementation Details}
\label{app:implementation}

\subsection{Models and context}
The default system uses the UrbanNav navigation policy~\citep{Mei_2026}, Qwen3.8-27B~\citep{qwen2026qwen38}, and the official pretrained Cosmos3-Nano checkpoint~\citep{nvidia2026cosmos3nano}. Pretrained models receive no weight updates during evaluation. Structured prompts implement action review, revision, visual comparison, and memory update. The goal description refers to the destination view; spatial phrases retain that reference frame.

\paragraph{Subgoal state.}
In the compact representation $S_t=(z_t,c_t,q_t,\sigma_t)$, the objective $z_t$ groups the semantic route intent and intermediate objective; cues $c_t$ group the visual anchor and progress or completion evidence. The loop serializes these as \texttt{intent}, \texttt{objective}, \texttt{anchor}, and \texttt{completion}; archived adapters also use the aliases \texttt{subgoal} and \texttt{completion\_evidence}. The open question and semantic status remain explicit. Creation step, allowed horizon, observed transition count, and accumulated movement are runtime metadata maintained by the harness. Expiration requests a fresh objective without asserting semantic failure. The latest assessment belongs to the associated experience record and may also be cached with the active context. Reasoning indices and rationales remain in the decision log rather than the core memory definition. The final destination remains the input to the local policy.

The preview--review--revise implementation retains four recent experience summaries and refreshes an active subgoal after six decisions if it has not already been achieved or contradicted. Its default inner-loop budget is $B=2$ reviewed actions.

\subsection{Action representation}
\label{app:action-representation}
UrbanNav produces eight local waypoints. The reasoner receives both their coordinates and a deterministic physical endpoint description, for example a displacement to the right and forward from the current camera. A constructed action uses cumulative ranges and bearings, which an adapter converts to waypoints. Invalid or low-confidence revisions fall back to the policy proposal. Historical action coordinates retain labels identifying their original source frame. NoMaD~\citep{sridhar2024nomad} and OmniVLA~\citep{hirose2026omnivla} use policy-specific action conversion and operating settings.

For a revised action, waypoint $k$ is specified by displacement $r_k$ from the current origin and bearing $\beta_k$ relative to forward. The conversion is $x_k=r_k\sin\beta_k$ and $y_k=r_k\cos\beta_k$, with distances in metres and angles in radians. The permitted action set is
\begin{equation}
 \mathcal A_{\mathrm{rev}}=
 \left\{a:
 0\leq r_k\leq r_{\max},\quad
 |\beta_k|\leq\beta_{\max},\quad
 y_K\geq y_{\min},\quad k=1,\ldots,K\right\}.
\label{eq:action-domain}
\end{equation}
The action interface supplies the range and bearing bounds and the minimum forward displacement. Validation also checks the output format, numerical values, confidence, and whether the revision repeats an earlier proposal. The frozen policy retains its original action interface; the revision prompt specifies its output requirements in Appendix~\ref{app:prompts}.

\subsection{Predictive Sandbox}
The implementation uses the Diffusers \texttt{Cosmos3OmniPipeline} and \texttt{CosmosActionCondition} in the \texttt{camera\_pose} domain. Each query receives the current image, a camera trajectory derived from the proposed movement, and a shared scene-continuation prompt, and returns 60 predicted frames in addition to the input frame. To construct the camera trajectory, we prepend the origin to the eight waypoints and linearly interpolate their positions to 61 poses along the full proposed path. Camera headings follow successive waypoint displacements and are interpolated after angle unwrapping, with the initial heading aligned with the current view. Adjacent poses yield 60 relative transformations, each encoded by three translation coordinates and a six-dimensional rotation representation. These indices parameterize the prediction, not physical execution time. Across revisions, the camera path changes while the image, text, and seed remain fixed. Generation uses seed 0, 30 diffusion steps, guidance 1.0, flow shift 10.0, the 480-resolution tier, ego view, no sound, and no system prompt.

Query records retain the source observation, proposed movement, route question, generated video, and action assessment, linking each predicted rollout to its conditioning action.

\subsection{UrbanNav evaluation}
The splits contain 4,179 test-seen and 4,634 test-unseen episodes. Each agent receives the destination description and egocentric observations and produces a sequence of local waypoints. All main comparisons use a maximum of 25 navigation decisions per episode and a three-metre success radius. Internal reasoning iterations and tool calls do not count as additional navigation decisions.

\begin{table}[t]
\centering
\caption{\textbf{Information exchanged through the harness.} Operations use the same vision--language reasoner. The table lists the inputs and outputs of each prompted operation.}
\label{tab:interfaces}
\small
\begin{tabularx}{\textwidth}{@{}l>{\raggedright\arraybackslash}X>{\raggedright\arraybackslash}X@{}}
\toprule
Operation & Principal inputs & Returned information\\
\midrule
Subgoal establishment & Destination, observed images, experience, previous objective & Route intent, anchor, objective, completion cue, open question.\\
Action Preview / Review & Current action, context, earlier reviews, optional predicted rollout & Action assessment, query usefulness, prediction quality, rationale, expected cues.\\
Action revision & Current action, its review, shared context, available images & Revised path or abstention, confidence, grounding, rationale, expected cues.\\
Memory Update & Subgoal, executed movement, expected cues, before/after real views & Observed-feedback verdict and rationale.\\
\bottomrule
\end{tabularx}
\end{table}

\subsection{CityWalker evaluation}
\label{app:citywalker}
The additional evaluation uses 154 route windows from nine recordings (IDs 10--18). Goals are fixed route endpoints expressed as rightward and forward displacements in the current frame. The frozen CityWalker policy~\citep{liu2025citywalker} receives five chronological RGB observations and coordinate context. All methods execute at most two metres along the proposed path per decision. Success requires reaching within three metres of the goal. Stationary transitions consume the 100-decision budget. Baselines use task-specific prompt adaptations. Path-length accounting is defined in Appendix~\ref{app:metrics}.

\section{Metrics and Complete Results}
\label{app:metrics}

We use the standard navigation measures SR, SPL, NE, and OSR~\citep{anderson2018navigationeval,anderson2018vln}. For episode $i$, let $S_i$ denote success, $d_i$ final distance to the goal, $l_i$ reference-route length, and $p_i$ executed waypoint-path length. Let $O_i$ indicate whether any executed waypoint enters the three-metre success radius. For $N$ episodes,
\begin{equation}
\mathrm{SR}=\frac{1}{N}\sum_i S_i,\quad
\mathrm{SPL}=\frac{1}{N}\sum_i S_i\frac{l_i}{\max(l_i,p_i)},\quad
\mathrm{NE}=\frac{1}{N}\sum_i d_i,\quad
\mathrm{OSR}=\frac{1}{N}\sum_i O_i.
\end{equation}
Trajectory alignment is measured by nDTW and SDTW~\citep{ilharco2019dtw}. For executed trajectory $\tau_i$ and reference trajectory $\tau_i^*$,
\begin{equation}
\mathrm{nDTW}_i=\exp\!\left(-\frac{\operatorname{DTW}(\tau_i,\tau_i^*)}{3|\tau_i^*|}\right),
\qquad
\mathrm{SDTW}=\frac{1}{N}\sum_i S_i\,\mathrm{nDTW}_i.
\end{equation}
We average nDTW across episodes. NE is measured in metres; other metrics are multiplied by 100. For SPL, $p_i$ sums the lengths of executed waypoint segments; $l_i$ is the reference-route length. Off-route termination counts as failure before metric aggregation.

\subsection{Full-split reporting}
\label{app:full-results}
Tables~\ref{tab:main}--\ref{tab:ablation} report all six metrics for the main comparison, policy transfer, and component ablation. Each row comes from a complete evaluation on its split. Table~\ref{tab:main} separates test-seen and test-unseen; Tables~\ref{tab:transfer} and~\ref{tab:ablation} use test-unseen.

\section{Supplementary Analyses}
\label{app:analysis}

\paragraph{Distribution-based curves.}
Figure~\ref{fig:route-gains} uses reference-length intervals with edges 0, 5, 10, 15, 20, 30, and 50 m, followed by an open-ended interval. Their episode counts are 517, 1,416, 1,086, 638, 540, 250, and 187. The turning intervals have edges 0, 90, 180, 270, 360, and 540 degrees, followed by an open-ended interval, with 922, 1,505, 1,210, 542, 303, and 152 episodes. Intervals include their lower edge and exclude their upper edge; each partition retains all 4,634 episodes. SR gains subtract the original policy's SR, whereas NE reductions subtract the method's NE from the policy's NE. The figure shows point estimates. Paired source-video-cluster bootstrap estimates use the protocol below and are retained with the machine-readable figure data. The subsequent binary-group analysis uses a coarser partition of the same route geometry.

\subsection{Route geometry}
The reference-length and cumulative-turning partitions are independent. Short routes have reference length $\leq20$ m (3,657 episodes), and long routes have length $>20$ m (977). Turning is the sum of absolute heading changes after retaining positions separated by at least 0.5 m. Merging the lower two turning tertiles produces groups of 3,089 and 1,545 episodes at $223.2241^\circ$. All methods use the same groups, determined solely by reference geometry. Table~\ref{tab:route-groups} uses the exact episode records for the harness with the Navigation Memory Tool and the full method in Table~\ref{tab:ablation}. Aggregating either partition recovers their overall SRs of 82.37\% and 83.38\%, and SPLs of 76.91\% and 77.87\%. Bootstrap sampling uses 33 source-video clusters, 10,000 replicates, and seed 20260915; intervals are conditional on the evaluated settings.

\input{provenance/generated_difficulty_intervals.tex}
\begin{table}[t]
\centering
\caption{\textbf{Route-group results.} SR/SPL show harness with Navigation Memory Tool / full method from Table~\ref{tab:ablation}. Each partition covers all 4,634 episodes. Intervals describe the paired gain in percentage points.}
\label{tab:route-groups}
\small
\setlength{\tabcolsep}{3pt}
\begin{tabularx}{\textwidth}{@{}Xrrrrr@{}}
\toprule
Group & Episodes & SR & $\Delta$SR interval & SPL & $\Delta$SPL interval\\
\midrule
\generateddifficultyintervals
\bottomrule
\end{tabularx}
\end{table}

\subsection{Paired memory and prediction probes}
\input{provenance/table_paired}

Table~\ref{tab:paired} summarizes changes to reasoning outputs when memory or predicted-video inputs change. Updating the subgoal context changes 199 of 1,501 policy-action assessments (13.3\%) and 134 retain/revise requests (8.9\%). In the preview-before-revision probe, revision requests increase from 147 to 214; among the 133 states where both conditions revise, the mean endpoint difference is 0.28 m. These measurements describe the reasoner's sensitivity to its inputs, rather than independent accuracy estimates.

At each of 525 fixed states, the construction probe supplies the same destination, observations, memory, and policy action, with or without its predicted rollout. Each condition constructs a new movement independently. These state-level probes are analysed separately from the full-episode navigation results.

The memory analysis includes 1,501 events at which an observed transition produces a new semantic assessment of an active subgoal context. Inherited assessments and expiration alone are excluded. Updates are supported in 633 events, achieved in 283, contradicted in 77, and unresolved in 508. At each event, paired review calls share current images, the policy proposal, observed memory, and recent heading, but receive the previous or updated subgoal context. Achieving an intermediate expectation is distinct from arriving at the final destination.

All probes use frozen Qwen with greedy decoding. Action construction allows 448 output tokens, memory-based review 64, and observed-cue verification 144. Predictions are sampled at 20-frame intervals from existing seed-0 videos exported at 10 fps. All construction, memory-review, and cue-verification calls yield valid structured outputs.

\begin{table}[t]
\centering
\caption{\textbf{Paired reasoning measurements.} Rates are percentages; rate differences are percentage points. The last row is in metres.}
\small
\begin{tabularx}{\textwidth}{@{}Xrrr@{}}
\toprule
Measurement & Pairs & Estimate & 95\% interval\\
\midrule
Memory: changed policy-action assessment & 1,501 & 13.26 & $[11.90,14.70]$\\
Memory: changed retain/revise request & 1,501 & 8.93 & $[7.62,10.18]$\\
Revision rate without prediction & 525 & 28.00 & $[21.56,35.44]$\\
Revision rate with prediction & 525 & 40.76 & $[34.63,47.30]$\\
Paired revision-rate difference & 525 & 12.76 & $[8.90,16.80]$\\
Endpoint change when both revise & 133 & 0.278 & $[0.215,0.342]$\\
\bottomrule
\end{tabularx}
\end{table}

Construction calls revise in both conditions at 133 states and retain the policy in both at 297; prediction introduces 81 revisions and withdraws 14. A predictive cue appears explicitly in 45 construction responses; the remaining 480 cite observed evidence as their basis.

\subsection{Observed feedback}
Stored action-selection records authorize the alternative at 366 paired states. The complete rollout has 358 matched executions, and 356 of these have an expected observation and a following observed view. The cue-verification prompt receives ordered before/after observed images, without generated frames, reference futures, or success labels. Frozen Qwen judges 183 cues supported (51.4\%), 39 contradicted (11.0\%), and 134 unresolved (37.6\%). These labels are model assessments, not independently annotated ground truth.

\section{Inference Cost}
\label{app:cost}

\paragraph{Experimental setup.}
We profile the predictive sandbox and the vision--language reasoner of \method{} separately on a single NVIDIA H200 GPU using BF16 inference. The reasoner is Qwen3.8-27B with FlashAttention-2, greedy decoding, a batch size of one, and a maximum output length of 384 tokens. Requests use the prompt templates provided in Appendix~\ref{app:prompts}, following one warm-up request per operation. Latency includes image loading, preprocessing, model inference, and response decoding, with GPU synchronization at the measurement boundaries. Model initialization and warm-up are excluded.

\paragraph{Predictive sandbox.}
Cosmos3-Nano generates action-conditioned rollouts with 60 predicted frames, using the 480-resolution setting, 30 denoising steps, and a guidance scale of 1.0. Table~\ref{tab:sandbox-cost} summarizes all 1,700 rollouts from the fastest complete generation run. Generation takes 11.37 s on average and uses 34.06 GiB of peak allocated GPU memory; timing includes video export.

\begin{table}[t]
\centering
\caption{\textbf{Predictive sandbox cost on one NVIDIA H200 GPU.} Statistics cover 1,700 rollouts, each containing 60 predicted frames. P90 denotes the 90th percentile.}
\label{tab:sandbox-cost}
\small
\begin{tabular}{@{}rrrr@{}}
\toprule
Mean (s) & Median (s) & P90 (s) & Peak GPU memory (GiB)\\
\midrule
11.37 & 11.35 & 11.47 & 34.06\\
\bottomrule
\end{tabular}
\end{table}

\paragraph{Reasoning harness.}
Table~\ref{tab:reasoner-cost} profiles VLM requests within the four-stage loop: subgoal establishment belongs to Navigation Memory, and revision is a separate request within Action Review. Subgoal establishment, Action Preview, and Action Review share the same states, with freshly established subgoals conditioning preview and review. Action Review receives four predicted frames in addition to three observed images. The revision measurement covers observation-based contradicted reviews; Memory Update uses observed before--after transitions. All measured requests, including abstentions, enter the statistics.

\begin{table}[t]
\centering
\caption{\textbf{Per-request latency on one NVIDIA H200 GPU at batch size one.} Predicted inputs use frame indices 10, 20, 40, and 60, with index 0 denoting the input observation. Each row times only the named operation; video generation is reported in Table~\ref{tab:sandbox-cost}.}
\label{tab:reasoner-cost}
\small
\setlength{\tabcolsep}{5pt}
\begin{tabularx}{\textwidth}{@{}Xrrrr@{}}
\toprule
Operation & Observed / predicted frames & Mean (s) & Median (s) & P90 (s)\\
\midrule
Subgoal establishment & 3 / 0 & 5.76 & 5.21 & 6.49\\
Action Preview & 3 / 0 & 5.06 & 5.04 & 5.41\\
Action Review & 3 / 4 & 5.19 & 5.05 & 5.65\\
Action Revision & 3 / 0 & 8.65 & 9.82 & 10.77\\
Memory Update & 2 / 0 & 2.57 & 2.35 & 2.91\\
\bottomrule
\end{tabularx}
\end{table}

The reasoner peaks at 51.89 GiB of allocated GPU memory across measured operations. Action Execute adds no VLM call; prediction and revision are conditional, with revised actions returning to preview and review.

%% file: appendix_prompts.tex
\section{Full Prompt Templates}
\label{app:prompts}

This section specifies the prompts for the reasoning loop in Section~\ref{sec:decision}: establishing a subgoal, reviewing a proposed action with or without prediction, constructing a revision and updating memory. Templates specify role instructions and response formats; double braces denote runtime substitutions. Line wrapping and JSON indentation are adjusted for readability. Case responses in Figure~\ref{fig:traces} instantiate these operations for illustration.

\subsection{Message construction}
\label{app:prompt-context}

\paragraph{Role and visual ordering.}
Each operation is sent as one user message containing its textual instructions and structured context, followed by real images in chronological order; the last real image is the current observation. These calls do not add a method-specific system message. For predicted review and revision, timestamped predicted frames are appended after the real images and explicitly labeled as predictions. Memory Update receives exactly two real images, before and after execution. The image placeholders below represent visual attachments, not strings sent in place of images.

\paragraph{Shared context.}
The destination text describes the view at the destination, not directions relative to the current agent. The current action is an $8\times2$ array in metres, with positive $x$ rightward and positive $y$ forward. Its endpoint description is computed from the last waypoint, not inferred from pixels. Earlier actions retain the coordinate frame in which they were proposed. Within a decision, observations, experience, and the subgoal stay fixed; only the current action and its review history change.

\begin{table}[t]
\centering
\caption{\textbf{Runtime substitutions in the prompt templates.} Placeholders are replaced by typed values before serialization; they are not additional model instructions.}
\label{tab:prompt-context}
\small
\renewcommand{\arraystretch}{1.12}
\begin{tabularx}{\textwidth}{@{}p{.29\textwidth}>{\raggedright\arraybackslash}X@{}}
\toprule
Placeholder & Value supplied to the operation\\
\midrule
\texttt{destination\_view} & The unmodified destination-view description $g$.\\
\texttt{observed\_experience} & Recent executed-action records: source step, action in its source frame, measured progress, expected cues, and observed feedback.\\
\texttt{subgoal} & Current objective, visual anchor, completion cue, open question, semantic status, and runtime metadata; \texttt{null} before initialization.\\
\texttt{current\_action\_m} & The proposed waypoint sequence, expressed in the current frame.\\
\texttt{endpoint\_x\_m}, \texttt{endpoint\_y\_m} & Coordinates of the final waypoint, formatted to three decimal places.\\
\texttt{earlier\_reviews} & Ordered action indices, waypoint sequences, and structured reviews; an empty list at the start of a decision.\\
\texttt{action\_assessment} & The structured review identifying the contradiction that a revision should address.\\
\texttt{executed\_action\_m}, \texttt{expected\_future\_cues} & The movement actually executed and the cues retained for checking the next real observation.\\
\bottomrule
\end{tabularx}
\end{table}

\paragraph{Structured responses.}
Vertical bars in the printed response formats enumerate valid alternatives; a response returns one value, not the entire list. A predicted review returns the same schema as an observation-based review. Action execution, waypoint decoding, tool dispatch, experience storage, and status transitions are harness operations rather than extra VLM calls.

The following schema expands the dynamic context. The single records shown stand for variable-length lists; the initial experience and review history are empty. A missing subgoal is serialized as \texttt{null}, and \texttt{latest\_feedback} is absent until a real transition has been assessed. The schema documents data construction, not an additional instruction appended to every request.

\begin{navpromptbox}{Shared context and within-decision review history}
\begin{lstlisting}[style=navprompt]
context = {
  "destination_view": {{destination_view_string}},
  "observed_experience": [
    {
      "source_step": {{source_decision_index}},
      "action_in_source_camera_m": {{executed_waypoints_8_by_2}},
      "progress_m": {{measured_real_progress}},
      "expected": {{expected_cue_string}},
      "observed_feedback": {
        "verdict": {{feedback_enum}},
        "reason": {{observed_evidence_string}}
      }
    }
  ],
  "subgoal": {
    "intent": {{route_intent_string}},
    "anchor": {{visible_anchor_string}},
    "objective": {{intermediate_objective_string}},
    "completion": {{observable_completion_cue_string}},
    "question": {{open_question_string}},
    "status": {{subgoal_status_enum}},
    "created_step": {{creation_decision_index}},
    "progress_m": {{accumulated_real_progress}},
    "latest_feedback": {
      "verdict": {{feedback_enum}},
      "reason": {{observed_evidence_string}}
    }
  }
}

earlier_reviews = [
  {
    "action_index": {{inner_loop_index}},
    "action_m": {{proposed_waypoints_8_by_2}},
    "review": {{complete_review_response}}
  }
]
\end{lstlisting}
\end{navpromptbox}

\subsection{Navigation Memory: establishing a subgoal}
\label{app:prompt-subgoal}

The harness issues this request when there is no active subgoal. The prompt receives the destination, retained experience, the previous objective if available, and observed images. The reasoner returns a route intent, visible anchor, intermediate objective, observable completion cue, and open question. The harness initializes \texttt{status=ACTIVE}; this status is not generated by the prompt.

\begin{navpromptbox}{Prompt 1. Establishing the current subgoal}
\begin{lstlisting}[style=navprompt]
Establish a tentative medium-horizon route objective for first-person navigation. Use a visible anchor and an observable completion cue; it may require several movements. Do not claim hidden geometry or arrival. Destination text describes the destination view. Output JSON only, each field under 16 words: 
{
  "intent": "route intent",
  "anchor": "visible anchor",
  "objective": "intermediate objective",
  "completion": "observable completion cue",
  "question": "unresolved question or NONE"
}.
{
  "destination_view": {{destination_view}},
  "observed_experience": {{observed_experience}},
  "previous_objective": {{subgoal}}
}
\end{lstlisting}
\end{navpromptbox}

The compact memory in Equation~\ref{eq:memory} groups \texttt{intent} and \texttt{objective} as \emph{Objective}, and \texttt{anchor} and \texttt{completion} as \emph{Visual cues}. The serialized \texttt{question} supplies \emph{Open question}. Creation step and accumulated progress are maintained outside this response.

\subsection{Action Preview: observation-based review}
\label{app:prompt-preview}

The first review examines one proposed movement against real observations and Navigation Memory. It returns both an action assessment and whether a prediction could answer a specific unresolved question. The harness calls the sandbox only for \texttt{UNRESOLVED} with \texttt{query\_useful=true}, as in Equation~\ref{eq:query}. A visible destination is not required for a local movement to be supported.

\begin{navpromptbox}{Prompt 2. Reviewing the proposed action without a rollout}
\begin{lstlisting}[style=navprompt]
Review the CURRENT proposed movement, not a tournament of candidates. Ordered factual images end with the current view. All action coordinates are relative to that current camera: x right, y forward, metres. Destination text describes the destination view. An unseen destination alone is not a reason to reject local movement. SUPPORTED requires visible or credible predicted route support. CONTRADICTED requires a concrete conflict. Missing evidence alone is UNRESOLVED. Prediction is useful only for a specific unresolved question about this movement. Generated views are fallible predictions, not factual observations or arrival evidence. Reject distorted or action-inconsistent video cues. No movement occurred between inner iterations. Current endpoint: right {{endpoint_x_m}} metres, forward {{endpoint_y_m}} metres.
{
  "context": {
    "destination_view": {{destination_view}},
    "observed_experience": {{observed_experience}},
    "subgoal": {{subgoal}}
  },
  "current_action_m": {{current_action_m}},
  "earlier_reviews": {{earlier_reviews}}
}
No forecast is supplied. A useful request must identify what the predicted view could clarify.
Output compact JSON only, text fields under 16 words: 
{
  "verdict": "SUPPORTED|CONTRADICTED|UNRESOLVED",
  "query_useful": false,
  "question": "specific question or NONE",
  "prediction_quality": "NOT_PROVIDED|COHERENT|INCONCLUSIVE|INCONSISTENT",
  "reason": "evidence",
  "expected": "cue to inspect after real execution"
}.
\end{lstlisting}
\end{navpromptbox}

\subsection{Action Review: predicted visual evidence}
\label{app:prompt-review}

After a sandbox query, the reasoner receives the same action, observations, memory, and review history, together with the predicted frames. The complete prompt is repeated here to make the two message variants independently usable. Its final instruction states that a forecast of this exact movement is already available, so the same action is not previewed again.

\begin{navpromptbox}{Prompt 3. Reviewing the proposed action with a rollout}
\begin{lstlisting}[style=navprompt]
Review the CURRENT proposed movement, not a tournament of candidates. Ordered factual images end with the current view. All action coordinates are relative to that current camera: x right, y forward, metres. Destination text describes the destination view. An unseen destination alone is not a reason to reject local movement. SUPPORTED requires visible or credible predicted route support. CONTRADICTED requires a concrete conflict. Missing evidence alone is UNRESOLVED. Prediction is useful only for a specific unresolved question about this movement. Generated views are fallible predictions, not factual observations or arrival evidence. Reject distorted or action-inconsistent video cues. No movement occurred between inner iterations. Current endpoint: right {{endpoint_x_m}} metres, forward {{endpoint_y_m}} metres.
{
  "context": {
    "destination_view": {{destination_view}},
    "observed_experience": {{observed_experience}},
    "subgoal": {{subgoal}}
  },
  "current_action_m": {{current_action_m}},
  "earlier_reviews": {{earlier_reviews}}
}
A forecast of this exact movement follows; do not request another forecast.
Output compact JSON only, text fields under 16 words: 
{
  "verdict": "SUPPORTED|CONTRADICTED|UNRESOLVED",
  "query_useful": false,
  "question": "specific question or NONE",
  "prediction_quality": "NOT_PROVIDED|COHERENT|INCONCLUSIVE|INCONSISTENT",
  "reason": "evidence",
  "expected": "cue to inspect after real execution"
}.
\end{lstlisting}
\end{navpromptbox}

\paragraph{Interpreting the review.}
\texttt{verdict} assesses the proposed action, whereas \texttt{prediction\_quality} assesses the supplied video. A coherent video need not support the action, and an inconclusive video does not itself contradict it. \texttt{reason} gives the short, evidence-based rationale; \texttt{expected} records the cue to inspect after real execution. With no video, prediction quality is \texttt{NOT\_PROVIDED}.

\subsection{Action Review: constructing a revision}
\label{app:prompt-revision}

A contradicted review and remaining budget permit a separate revision request. This call receives the current action, its assessment, earlier reviews, and the same observed and predicted images available to the review. It preserves the current subgoal while changing the local movement.

\begin{navpromptbox}{Prompt 4. Revising a contradicted movement}
\begin{lstlisting}[style=navprompt]
Revise the current movement to address its identified contradiction. Use current observed walkable space; predicted frames can explain the problem but do not establish hidden free space. Preserve the medium-horizon objective. No movement has occurred during this reasoning loop. Output eight [range_cm,bearing_deg] positions at one-second intervals, all relative to the CURRENT camera origin. Range is nonnegative total displacement in centimetres, not a step increment; bearing -90..90 degrees (negative left, positive right). Choose a movement distance supported by the currently visible walkable space. The last position must be at least 10 cm forward. Do not repeat an earlier action. The new action will be reviewed before execution. If no grounded revision exists, ABSTAIN.
{
  "context": {
    "destination_view": {{destination_view}},
    "observed_experience": {{observed_experience}},
    "subgoal": {{subgoal}}
  },
  "current_action_m": {{current_action_m}},
  "assessment": {{action_assessment}},
  "earlier_reviews": {{earlier_reviews}}
}
Output JSON only, text fields under 16 words: 
{
  "decision": "REVISE|ABSTAIN",
  "confidence": "LOW|MEDIUM|HIGH",
  "path": [],
  "grounding": "current visible free space",
  "reason": "how the change addresses the issue",
  "expected": "observable next-view cue"
}.
\end{lstlisting}
\end{navpromptbox}

\paragraph{Decoding and re-review.}
On \texttt{REVISE}, \texttt{path} must contain eight numeric pairs; the empty array in the printed format is a schema placeholder, not a valid revised path. Each pair specifies range from the current origin in centimetres and bearing in degrees. Appendix~\ref{app:action-representation} describes the coordinate conversion; Equation~\ref{eq:action-domain} specifies the bounds. \texttt{ABSTAIN} requires an empty path. The harness checks the decoded action, rejects low-confidence or repeated revisions, and sends a valid new movement back to Action Preview. The review and revision outputs are displayed together in Figure~\ref{fig:traces}, but they come from distinct operations.

\subsection{Memory Update: execution feedback}
\label{app:prompt-update}

After Action Execute, the reasoner compares two real images with the executed movement, current subgoal, and retained expected cues. Predicted frames are not attached to this request. The resulting \texttt{verdict} is observed feedback, not another pre-execution action assessment.

\begin{navpromptbox}{Prompt 5. Assessing the observed transition}
\begin{lstlisting}[style=navprompt]
Assess the actual transition against the stored medium-horizon objective. First image is BEFORE execution, second AFTER. SUPPORTED means compatible progress, not completion; ACHIEVED requires the completion cue to be visible. CONTRADICTED requires conflicting observed evidence; otherwise UNRESOLVED. Historical action coordinates belong to the BEFORE view. Return JSON 
{
  "verdict": "SUPPORTED|ACHIEVED|CONTRADICTED|UNRESOLVED",
  "reason": "specific observed cue, under 20 words"
}.
{
  "destination_view": {{destination_view}},
  "subgoal": {{subgoal}},
  "executed_action_m": {{executed_action_m}},
  "expected": {{expected_future_cues}}
}
\end{lstlisting}
\end{navpromptbox}

\begin{table}[t]
\centering
\caption{\textbf{From observed feedback to memory state.} Feedback is generated by the VLM; storage and semantic-status updates are performed by the harness.}
\label{tab:memory-transitions}
\small
\renewcommand{\arraystretch}{1.12}
\begin{tabularx}{\textwidth}{@{}ll>{\raggedright\arraybackslash}X@{}}
\toprule
Observed feedback & Subgoal status & Next operation\\
\midrule
\texttt{SUPPORTED} & \texttt{ACTIVE} & Retain the objective and append evidence of compatible progress.\\
\texttt{UNRESOLVED} & \texttt{ACTIVE} & Retain the objective and unresolved question.\\
\texttt{ACHIEVED} & \texttt{ACHIEVED} & Close the completed subgoal; establish the next objective.\\
\texttt{CONTRADICTED} & \texttt{CONTRADICTED} & Reconsider the objective using the conflicting observation.\\
\bottomrule
\end{tabularx}
\end{table}

Every real transition is stored with its feedback. A context-lifetime limit can additionally expire an active subgoal and request a refresh; expiration is runtime bookkeeping, not a claim of semantic failure. The figure's \emph{Open question progress} summarizes the feedback rationale rather than an independently predicted JSON field. Likewise, achieving a subgoal does not authorize stopping at the final destination.

\subsection{Predictive Sandbox: world-model request}
\label{app:prompt-sandbox}

The world model receives the current image, the proposed movement through the action adapter, and the following fixed scene-continuation text. It does not receive the destination description, reference future, success label, or subgoal as generation text. The subgoal and open question instead guide the VLM's interpretation of the returned video.

\begin{navpromptbox}{Prompt 6. Scene continuation for the action-conditioned world model}
\begin{lstlisting}[style=navprompt]
A continuous realistic first-person video of the same urban sidewalk scene. Preserve the visible buildings, vehicles, pedestrians, obstacles, and lighting. The camera motion follows the supplied camera trajectory.
\end{lstlisting}
\end{navpromptbox}

The request uses an empty negative prompt and no added system prompt. Across revisions at the same real observation, the image, scene text, and seed remain fixed while the movement condition changes. Generation settings and action encoding are given in Appendix~\ref{app:implementation}.

\subsection{Correspondence with the method and figures}
\label{app:prompt-mapping}

\begin{table}[t]
\centering
\caption{\textbf{Prompt outputs in the method and qualitative figures.} Descriptive figure labels expose the semantics of the serialized fields without adding new model outputs.}
\label{tab:prompt-fields}
\small
\renewcommand{\arraystretch}{1.12}
\begin{tabularx}{\textwidth}{@{}p{.29\textwidth}p{.27\textwidth}>{\raggedright\arraybackslash}X@{}}
\toprule
Displayed field & Runtime field & Role in the method\\
\midrule
Objective & \texttt{intent}, \texttt{objective} & Medium-horizon intent $z_t$ in Equation~\ref{eq:memory}.\\
Visual cues & \texttt{anchor}, \texttt{completion} & Observable anchors and progress/completion cues $c_t$.\\
Open question & \texttt{question} & Unresolved question $q_t$ in memory; review can specify a query about the current action.\\
Action assessment & review \texttt{verdict} & $\omega_t^j$ in Equation~\ref{eq:assessment}.\\
Preview useful & \texttt{query\_useful} & $\kappa_t^j$ in Equation~\ref{eq:assessment}.\\
Prediction quality & \texttt{prediction\_quality} & Video assessment $\chi_t^j$.\\
Rationale & \texttt{reason} & Short explanation $\rho_t^j$, or the explanation accompanying observed feedback.\\
Expected future cues & \texttt{expected} & $\nu_t^j$, retained for the next real transition.\\
Revised action & decoded \texttt{path} & $a_t^{j+1}$; the figure shows its endpoint rather than all eight waypoints.\\
Observed feedback & feedback \texttt{verdict} & $v_{t+1}$ in Equation~\ref{eq:feedback}.\\
Status & harness-maintained \texttt{status} & $\sigma_t$; updated according to Table~\ref{tab:memory-transitions}.\\
Open question progress & summary of feedback & A readable description of progress, not a separate response key.\\
Observed experience & harness transition record & Append $(o_t,a_t,o_{t+1},v_{t+1})$ after real execution.\\
\bottomrule
\end{tabularx}
\end{table}

Figure~\ref{fig:architecture} uses a compact four-field subgoal; Figure~\ref{fig:traces} expands the corresponding prompt inputs and outputs. In the parked-vehicle case, Action Preview requests missing evidence, the predicted review identifies the storefront-facing movement, and Revision constructs a sidewalk-following movement for re-review. In the pedestrian case, contradictory real feedback closes the earlier subgoal before a new one is established. The same templates support both transitions; case-specific statements fill the context and response fields rather than change the protocol.

%% file: provenance/generated_difficulty_intervals.tex
\newcommand{\generateddifficultyintervals}{%
Short & 3,657 & 94.26 / 94.56 & $[0.10,0.51]$ & 87.65 / 87.91 & $[0.08,0.47]$\\
Long & 977 & 37.87 / 41.56 & $[1.82,5.57]$ & 36.71 / 40.26 & $[1.72,5.39]$\\
Low + medium turn & 3,089 & 94.92 / 95.40 & $[0.27,0.72]$ & 88.48 / 88.92 & $[0.24,0.66]$\\
High turn & 1,545 & 57.28 / 59.35 & $[1.04,3.07]$ & 53.77 / 55.77 & $[0.98,2.98]$\\
}

%% file: provenance/table_paired.tex
\begin{table}[t]
\centering
\caption{\textbf{Paired effects on reasoning.} Each comparison holds observed images, destination, and policy proposal fixed. Only the memory update or predicted-video input changes. These are state-level probes, separate from the navigation rollouts.}
\label{tab:paired}
\navtableformat
\begin{tabularx}{\textwidth}{@{}lXrr@{}}
\toprule
Changed input & Measurement & Pairs & Result\\
\midrule
Updated memory & Policy-action assessment changed & 1,501 & 13.3\%\\
Updated memory & Retain/revise decision changed & 1,501 & 8.9\%\\
Predicted video & States requesting a revision & 525 & $147\to214$\\
Predicted video & Endpoint change when both revise & 133 & 0.28 m\\
\bottomrule
\end{tabularx}
\end{table}